\documentclass[runningheads]{llncs}
\usepackage[T1]{fontenc}
\usepackage{graphicx}
\usepackage{amsmath}
\usepackage{amssymb}
\usepackage{booktabs}
\usepackage{multirow}
\usepackage{bbding}

\begin{document}

\title{Degraded Infrared Small Object Detection\\via Degradation-Adapted Physics-Guided Restoration}
\titlerunning{Degraded Infrared Small Object Detection}

\author{Xinkai Lu\inst{1,3} \and Wenjun Chen\inst{1} \and
Yi Li\inst{2}\thanks{Corresponding author.} \and Yi Chang\inst{1} \and
Luxin Yan\inst{1}}
\authorrunning{X. Lu et al.}
\institute{National Key Lab of Multispectral Information Intelligent Processing Technology,\\
School of Artificial Intelligence and Automation,\\
Huazhong University of Science and Technology, Wuhan, China
\and
School of Ocean Engineering, Tsinghua Shenzhen International Graduate School,\\
Tsinghua University, Shenzhen, China
\and
Yichang Testing Technique Research Institute, Yichang, China\\
\email{luxinkai6573@163.com, \{wjchen, yichang, yanluxin\}@hust.edu.cn, li\_yi@tsinghua.edu.cn}}

\maketitle

\begin{abstract}
Infrared small object detection has made significant progress in recent years. However, degradations such as fog and nonuniformity can suppress target-background contrast, substantially increasing detection difficulty. Existing methods mainly rely on image restoration as preprocessing, but they are typically designed for specific degradation types and fail to generalize to varying degradations. To alleviate this, we propose DAISOD, a degradation-adapted infrared small object detection framework for robust detection under different degradations. DAISOD first identifies the type and severity of degradations, then adapts the processing via dedicated branches, and finally fuses the results for subsequent detection. Moreover, a physics-guided restoration mechanism is incorporated to explicitly estimate degradation parameters and remove degradation effects through physical models, avoiding excessive restoration that may erase small targets. Moreover, we construct a degraded infrared small object detection dataset covering diverse degradation types and levels. Extensive experiments show that DAISOD outperforms state-of-the-art methods under various degradation conditions.

\end{abstract}

\section{Introduction}
\label{sec:intro}

Infrared small object detection (ISOD) plays a crucial role in unmanned surveillance and perception systems \cite{dai2021attentional,wu2022uiu} and has attracted extensive research attention. However, degradations such as fog and nonuniformity are ubiquitous in real-world applications, significantly deteriorating infrared image quality and posing severe challenges to small object detection. Fog-induced atmospheric scattering \cite{narasimhan2002vision,li2017aod} and nonuniform background interference both lead to pronounced contrast degradation, which severely suppresses the already limited target-background separability in infrared imagery. As a result, reliable feature extraction becomes more difficult, causing substantial performance degradation in ISOD.

\begin{figure}[!t]
  \centering
  \includegraphics[width=\textwidth]{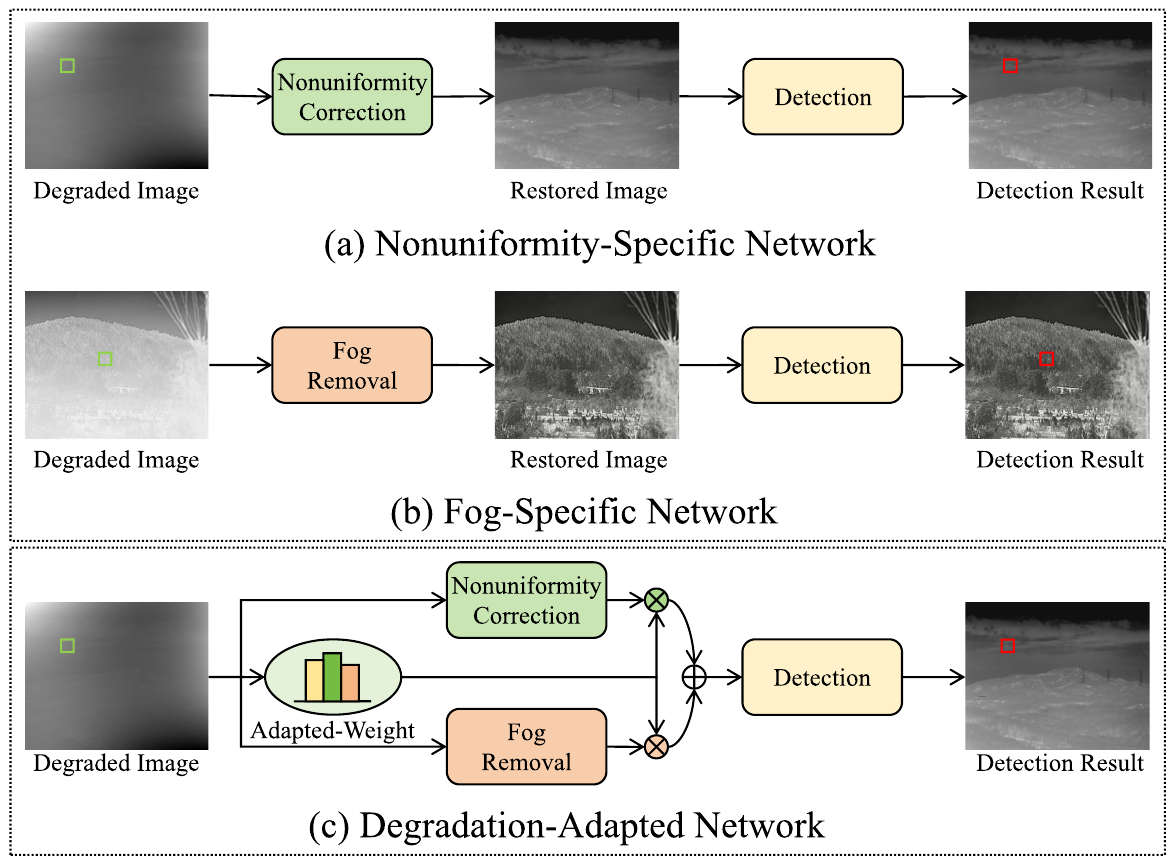}
  \caption{Comparison of detection pipelines under degraded conditions. 
(a)-(b) Fixed task-specific pipelines in existing methods limit generalization across diverse degradations.
(c) Our DAISOD dynamically adapts physics-guided branches through an adaptive weighting mechanism for robust detection.}

  \label{fig:motivation}
\end{figure}

To address detection under complex degradations, existing methods can be mainly classified into two categories: image-level methods \cite{qin2020ffa,li2022single,joint_restoration,wang2023dehazing,fu2025thermal} and feature-level methods \cite{ying2023mapping,liu2023infrared,domain-invariant,hu2025nighttime}. Image-level methods explicitly incorporate image restoration as a preprocessing step to remove degradation effects \cite{wang2023dehazing,fu2025thermal}. Although promising performance has been achieved, these methods are usually designed only for a specific type of degradation and thus struggle to generalize across different degradation scenarios \cite{zhang2025unsupervised,cheng2025hyperspectral}. More critically, due to their extremely small scale and weak signals, infrared small objects are statistically entangled with degradations, making direct end-to-end restoration prone to eroding vulnerable target signals \cite{wang2025nonuniform,hu2025nighttime}.

In contrast, feature-level approaches attempt to alleviate degradation effects by learning degradation-robust representations in the feature space \cite{li2022dense,hu2025nighttime,tang2025ffta}. However, infrared small object detection relies on inherently extremely subtle target-background distinctions \cite{itti2025bayesian,tang2025ffta}, which can be easily overwhelmed by strong degradations. Furthermore, the highly distinct characteristics of various degradations make it particularly challenging to learn unified and robust feature representations directly from degraded images, thereby limiting generalization across multiple conditions \cite{zhang2025unsupervised,cheng2025hyperspectral}.

The above observations highlight a central challenge in ISOD under complex environments: suppressing diverse degradations without damaging inherently weak target signals. To this end, we propose a degradation-adapted infrared small object detection framework, termed DAISOD, which performs degradation-specific restoration guided by corresponding physical models to precisely remove degradation interference while preserving target information \cite{wang2023dehazing,fu2025thermal}.

Specifically, the proposed framework first analyzes the degradation type and severity in the input infrared images and performs degradation-adapted, physics-guided image restoration to handle diverse degradation scenarios. The restoration network estimates key parameters of physical degradation models, enabling explicit and controllable removal of degradation effects while preserving small object information \cite{wang2023dehazing}. To further enhance detection robustness and generalization, a feature-level contrastive loss is incorporated to improve the discriminability of learned representations \cite{cheng2025hyperspectral}. The framework is optimized using a staged strategy: the restoration branches are first pretrained, after which the adaptive fusion head and detector are jointly optimized for ISOD under complex environments. In addition, a degraded infrared small object detection dataset with multiple degradation types and severity levels is constructed via physics-based simulation to support comprehensive evaluation. The main contributions are summarized as follows.

\begin{itemize}
    \item \noindent We address the challenge of diverse and uncertain degradations in infrared small object detection by proposing a unified degradation-adapted framework, termed DAISOD, which jointly handles fog and nonuniformity within a single model.
    \item \noindent We introduce a physics-guided degradation-adaptive restoration module that explicitly estimates physical degradation parameters and removes degradation effects through physical modeling, enabling effective suppression of degradations while preserving small object information.
    \item \noindent We construct a degraded infrared small object detection dataset via physics-based simulation with multiple degradation types and severity levels, on which extensive experiments demonstrate the superior robustness of DAISOD under diverse degradation conditions.
\end{itemize}

\begin{figure}[!t]
  \centering
  \includegraphics[width=\textwidth]{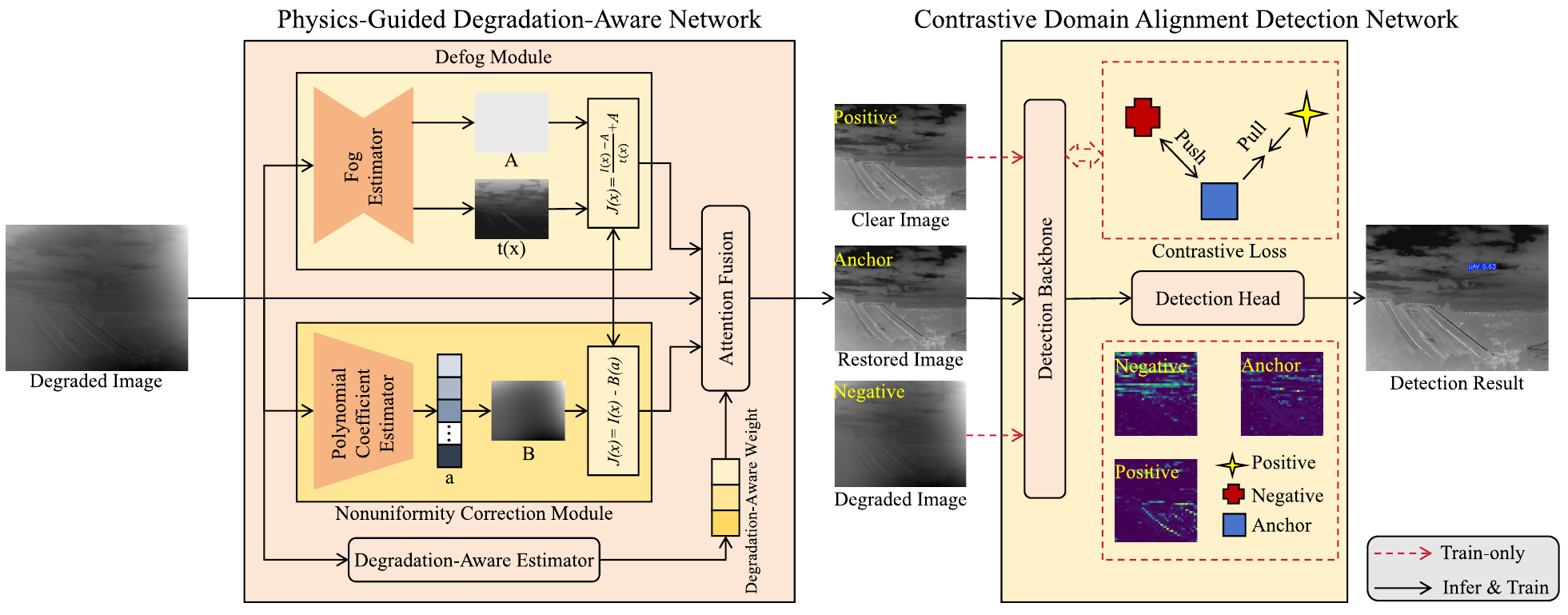}
  \caption{Architecture of the proposed DAISOD. 
  The framework integrates a physics-guided degradation-aware network (left) and a detection network (right). The former utilizes parallel estimators to infer physical parameters. Crucially, it employs a degradation-aware estimator to explicitly perceive the degradation context, dynamically modulating the contribution of each physics-guided branch for precise adaptation. The latter employs contrastive learning to align restored features with the clean domain while distancing them from the degraded domain, ensuring degradation-invariant representations.}
  \label{fig:architecture}
\end{figure}

\section{Methodology}
\label{sec:method}

\subsection{Overall Architecture}
In this work, we focus on the challenging problem of infrared small object detection under heterogeneous degradations. The crux lies in the statistical ambiguity between uncertain degradation patterns and vulnerable small targets, which causes conventional blind restoration methods to inadvertently erode target signals. To tackle this, we introduce DAISOD (Fig.~\ref{fig:architecture}), a framework designed to break this ambiguity via physics-guided constraints and feature alignment. Specifically, the physics-guided degradation-aware network explicitly estimates degradation parameters to prevent the network from misinterpreting targets as interference, while the contrastive domain alignment detection network enforces feature-level consistency with the clean domain, ensuring that the restored targets are statistically robust for detection.

\subsection{Physics-Guided Degradation-Aware Network}
\label{sec:pgaf}

Real-world infrared images suffer from coupled degradations. Blind end-to-end restoration risks eroding small targets by treating these as generic noise. To address this, we decouple restoration via physics-guided branches, transforming black-box mappings into transparent parameter estimation.

\textbf{Parametric Nonuniformity Correction.} 
Existing scene-based correction methods often rely on strong motion assumptions, failing in static scenarios, while pure end-to-end deep networks are prone to overfitting. To address this, we propose a hybrid correction scheme that embeds parametric priors into a lightweight deep network.
We model the degradation as an additive nonuniformity field $B(\mathbf{x})$, formulated as $I(\mathbf{x}) = J(\mathbf{x}) + B(\mathbf{x})$. Instead of predicting a pixel-wise map, we constrain $B(\mathbf{x})$ using a globally smooth bivariate polynomial surface. For a pixel at coordinate $\mathbf{x}=(x,y)$, the bias is defined as:
\begin{equation}
B(\mathbf{x}) = \sum_{i=0}^{D} \sum_{j=0}^{D-i} \alpha_{i,j} x^i y^j = \mathbf{p}(\mathbf{x})^\top \boldsymbol{\alpha}
\end{equation}
We employ a lightweight coefficient estimator (ResNet-18) to regress the global coefficient vector $\boldsymbol{\alpha}$. The corrected image is strictly obtained by $J_{nuc} = I - B(\mathbf{x})$, which effectively removes low-frequency bias while preserving high-frequency target structural details.

\textbf{Physics-Based Defogging.} 
Similarly, for atmospheric scattering, traditional priors are inapplicable to single-channel infrared images, while blind restoration networks often lack physical constraints. To ensure restoration fidelity, we adhere to the physical formation model governed by Koschmieder's law: $I(\mathbf{x}) = J(\mathbf{x})t(\mathbf{x}) + A(1-t(\mathbf{x}))$.
We employ a U-Net-based fog estimator to estimate the transmission map $t(\mathbf{x})$ and atmospheric light $A$. The fog-free component $J_{fog}$ is then physically recovered via the process of explicit inversion.
\begin{equation}
J_{fog}(\mathbf{x}) = \frac{I(\mathbf{x}) - A}{t(\mathbf{x})} + A
\end{equation}
 This explicit decoupling ensures that the restoration adheres to physical laws, avoiding intensity distortion.

\textbf{Degradation-Aware Adaptive Fusion.} 
Direct concatenation of multi-branch outputs inevitably introduces information redundancy and conflicting artifacts, as the network struggles to differentiate valid restoration from residual degradations across channels. To mitigate this, we propose a weighted concat fusion strategy. Specifically, the original input $I$, nonuniformity-corrected $J_{nuc}$, and defogged $J_{fog}$ are first modulated by the degradation-aware weights $\omega$ to suppress invalid branches explicitly. The weighted features are then concatenated and fused via a 3-layer convolutional network. This mechanism acts as a soft feature selection process, ensuring that only complementary, high-quality information is integrated into the final output $J$.

\subsection{Contrastive Domain Alignment and Optimization}

To bridge the domain gap between restored and clean features, we introduce a contrastive domain alignment regularization mechanism, which is jointly optimized with the reconstruction and detection objectives.

\textbf{Motivation and Contrastive Domain Alignment Construction.}
Unlike generic contrastive representation learning, our objective is not to calibrate an absolute feature distance or discriminate image identities. Feature distances are representation-dependent and may vary considerably across network layers and degradation conditions. Instead, we focus on the relative domain relationship: the restored representation should be closer to its clean counterpart than to the corresponding degraded observation.

For each training sample, we construct a content-aligned triplet in which the restored image $J$ is treated as the anchor, the clean
ground-truth image $J_{gt}$ as the positive, and the corresponding degraded input $I$ as the paired negative. Since these three images share the same scene content and small target, their relative feature distances primarily characterize degradation-induced discrepancies rather than unrelated semantic differences. Motivated by ratio-based contrastive regularization in image restoration
\cite{wu2021contrastive,zheng2023curricular}, we define the contrastive domain alignment loss as
\begin{equation}
	\mathcal{L}_{cda}
	=
	\sum_l \omega_l
	\frac{
		\left\| \phi_l(J)-\phi_l(J_{gt}) \right\|_1
	}{
		\left\| \phi_l(J)-\phi_l(I) \right\|_1+\epsilon
	},
	\label{eq:cda}
\end{equation}
where $\phi_l$ denotes the feature representation extracted from the $l$-th selected layer, $\omega_l$ is the corresponding layer weight, and $\epsilon$ is a small constant for numerical stability. The numerator measures the positive distance between the restored and clean representations, while the denominator measures the negative distance between the restored and degraded representations. Minimizing $\mathcal{L}_{cda}$ reduces the clean-domain distance relative to the degraded-domain distance, thereby encouraging the restored representation to be closer to the clean reference than to the degraded input.

\textbf{Comparison with Existing Contrastive Objectives.}
Different from existing contrastive objectives such as InfoNCE \cite{oord2018representation}, the adopted distance-ratio formulation relies on a content-consistent paired negative rather than instance discrimination over a large negative set. Applying InfoNCE to our setting would typically require additional in-batch or memory-bank negatives. Using infrared images from different scenes as such negatives may introduce variations in scene content, background structure, and target location, causing the learned relation to be dominated by semantic differences rather than degradation-induced discrepancies. Moreover, InfoNCE introduces additional dependence on the construction of the negative set and the selection of the temperature hyperparameter.

Moreover, compared with the standard triplet loss \cite{schroff2015facenet}, which imposes a fixed additive margin, the distance ratio provides a scale-adaptive relative constraint by normalizing the positive distance with the corresponding paired negative distance. This avoids applying a common absolute margin to heterogeneous feature layers and degradation conditions. In addition, the ratio-based objective generally remains active as long as the positive distance is nonzero, whereas the standard margin-based triplet loss becomes inactive once its prescribed margin is satisfied. Therefore, this formulation is selected because it is consistent with our paired restoration setting, rather than as a universally superior replacement for other contrastive objectives.

\subsection{Overall Loss and Implementation Details}

\textbf{Overall Loss Function.}
The contrastive term is used as a relative feature-space regularizer rather than an independent restoration objective. Specifically, the reconstruction loss provides absolute pixel-level fidelity, the contrastive domain alignment loss imposes relative feature-space discrimination, and the detection loss preserves task-relevant semantics. The complete training objective is
\begin{equation}
	\mathcal{L}_{total}
	=
	\lambda_{rec}\mathcal{L}_{rec}
	+
	\lambda_{cda}\mathcal{L}_{cda}
	+
	\lambda_{det}(t)\mathcal{L}_{det},
	\label{eq:total_loss}
\end{equation}
where $\mathcal{L}_{rec}$ denotes the pixel-wise reconstruction loss, and $\mathcal{L}_{det}$ follows the standard YOLO11n objective \cite{yolo11_ultralytics}, comprising CIoU, BCE, and DFL losses. In our implementation, we set $\lambda_{rec}=1.0$ and $\lambda_{cda}=0.1$. To stabilize multi-task learning, we employ a dynamic warmup strategy for $\lambda_{det}(t)$ ($0.001 \rightarrow 0.01$), prioritizing restoration convergence during the early training stage before imposing stronger detection constraints.

\textbf{Implementation Details.}
The degradation-aware estimator is implemented as a lightweight three-stage convolutional encoder. Each stage consists of a $3\times3$ convolution with a stride of 2, followed by batch normalization, ReLU activation, and a squeeze-and-excitation block for channel-wise feature recalibration. The channel dimensions of the three stages are set to 32, 64, and 128, respectively. The resulting features are aggregated by global average pooling and fed into a two-layer prediction head with dimensions $128\rightarrow64\rightarrow3$. A sigmoid function produces three image-level gating scores $\boldsymbol{\omega}=[\omega_{raw},\omega_{fog},\omega_{nuc}]$, which correspond to the original, defogging, and nonuniformity-correction branches, respectively. During restoration pretraining, the estimator is supervised by degradation-category labels to encourage the branch associated with the observed degradation to receive a stronger response.

For network compatibility, the single-channel infrared images are represented in three-channel format. Given the predicted gating scores, each scalar is spatially broadcast and multiplied by its corresponding three-channel representation. The weighted streams are then concatenated along the channel dimension as
$\mathbf{F}_{cat}=\operatorname{Concat}(\omega_{raw}I,
\omega_{fog}J_{fog},\omega_{nuc}J_{nuc})\in
\mathbb{R}^{9\times H\times W}$. Instead of directly summing the three branches, we feed $\mathbf{F}_{cat}$ into a lightweight fusion head consisting of two $3\times3$ convolutional layers with 32 channels, each followed by batch normalization and ReLU, and a final $1\times1$ convolution that maps the features back to three channels. A sigmoid function produces the final restored image $J$. This weighted-concat design preserves branch-specific information before nonlinear feature mixing, allowing the fusion head to suppress invalid restoration responses while retaining complementary details. During joint optimization, the pretrained restoration branches and degradation-aware estimator are fixed, whereas the fusion head is further refined using the reconstruction and downstream detection objectives.

\subsection{IR-HDSOD Dataset}
Existing benchmarks primarily focus on clean or single-degradation scenarios, failing to reflect the coupled interference ubiquitous in real-world surveillance. To bridge this gap, we construct IR-HDSOD, a large-scale benchmark containing 128,000 paired samples. 
Base images are aggregated from public datasets and self-collected footage across four distinct scenarios: sky, sea, urban, and forest. 
To simulate realistic degradation, we synthesize physically coupled interference by randomly combining three intensity levels of atmospheric fog and sensor nonuniformity.
The dataset features a total of 130,790 annotated bounding boxes, categorized into 120,582 planes and 10,208 ships. 
Unlike previous benchmarks, IR-HDSOD provides strictly coupled environmental and sensor degradations together with pixel-aligned clean references, covering extreme variations in target scales and background clutter to enable rigorous robustness evaluation.

\section{Experiments}
\label{sec:experiments}

\subsection{Experimental Settings}

\begin{figure}[t]
  \centering
  \includegraphics[width=\textwidth]{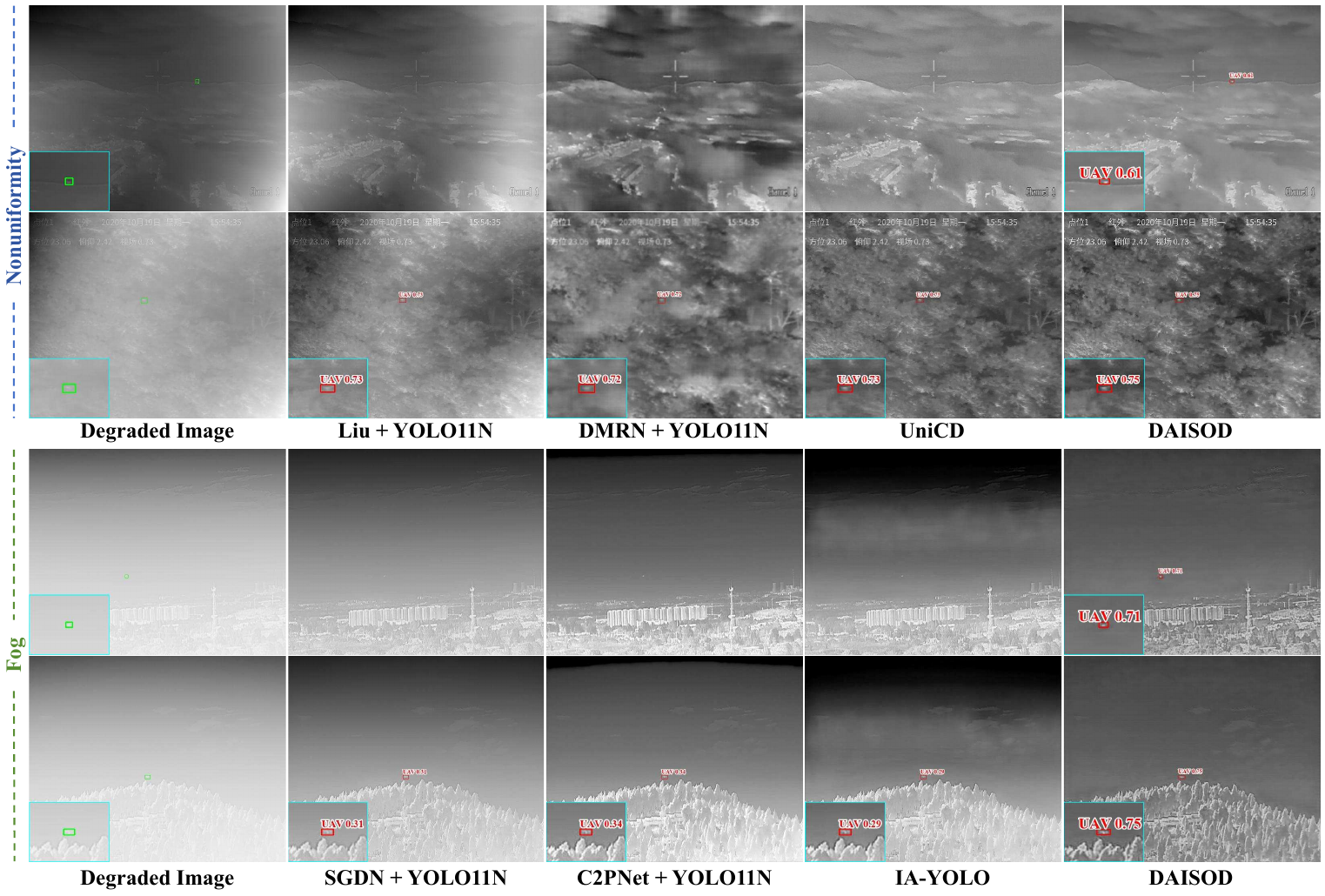}
  \caption{Visual comparison of detection results between SOTA methods and our DAISOD under degraded conditions. The top rows show results under nonuniformity conditions, while the bottom rows display results in foggy scenes. Green and red boxes represent ground-truth and detected targets, respectively.}
  \label{fig:detection_results}
\end{figure}

We evaluate DAISOD on the proposed IR-HDSOD dataset (split 8:1:1), utilizing mAP$_{50}$ for detection and PSNR/SSIM for restoration fidelity. Implemented in PyTorch 2.1 on an NVIDIA RTX 3090, the model is trained for 50 epochs (batch size 16) using the Adam optimizer (initial $lr=1\times10^{-3}$, cosine decay, weight decay $1\times10^{-4}$). Random horizontal flipping is applied for augmentation. For fair comparison, the detection backbone for all separate and direct baselines is standardized to YOLO11n.

\subsection{Comparison with State-of-the-Art Methods}

We benchmark DAISOD against three categories of approaches: 
(1) \textbf{Direct Detection}: Training YOLO11n \cite{yolo11_ultralytics} directly on degraded images.
(2) \textbf{Separate Methods}: Cascading SOTA restoration networks (e.g., FFA-Net \cite{qin2020ffa}, C2PNet \cite{zheng2023curricular}, SGDN \cite{fang2025guided}, Liu \cite{LIU2016235}, DMRN \cite{chang2019infrared}, AHBC \cite{ahbc}) with a fixed YOLO11n detector.
(3) \textbf{Unified/Adaptation Methods}: Utilizing recent domain adaptation (SF-UT \cite{hao_simplifying_2024}) or image-adaptive frameworks (IA-YOLO \cite{liu2022image}, UniCD \cite{fang2025detection}).

\begin{table}[t]
\centering
\caption{Quantitative comparison under Foggy conditions. The best results are highlighted in bold.}
\label{tab:fog_results}
\resizebox{\textwidth}{!}{
\begin{tabular}{lccccc}
\toprule
Strategy & Method & Pub. & PSNR $\uparrow$ & SSIM $\uparrow$ & mAP $\uparrow$ \\
\midrule
Direct & YOLO11n & - & 10.76 & 0.7272 & 30.44 \\
\midrule
\multirow{3}{*}{Separate} & FFA-Net \cite{qin2020ffa} + YOLO11n & AAAI'20 & 15.15 & 0.7265 & 29.37 \\
 & C2PNet \cite{zheng2023curricular} + YOLO11n & CVPR'23 & 19.37 & 0.8052 & 39.98 \\
 & SGDN \cite{fang2025guided} + YOLO11n & AAAI'25 & 14.87 & 0.7277 & 31.16 \\
\midrule
Domain Adapt & SF-UT \cite{hao_simplifying_2024} & ECCV'24 & - & - & 48.98 \\
Union & IA-YOLO \cite{liu2022image} & AAAI'22 & 14.83 & 0.7110 & 34.09 \\
\midrule
\textbf{Ours} & \textbf{DAISOD} & - & \textbf{19.95} & \textbf{0.8274} & \textbf{49.97} \\
\bottomrule
\end{tabular}
}
\end{table}

\textbf{Results under Foggy Conditions.} 
As shown in Table~\ref{tab:fog_results}, separate restoration-and-detection strategies such as FFA-Net and SGDN fail to improve upon the direct baseline. Notably, FFA-Net reduces mAP from 30.44\% to 29.37\%, suggesting that blind restoration may erode weak target signals. While the domain adaptation method SF-UT achieves a competitive mAP of 48.98\%, DAISOD outperforms it by 0.99 percentage points and substantially surpasses IA-YOLO. Together with the ablation results, these comparisons support the effectiveness of the proposed physics-guided restoration and contrastive domain alignment in preserving target features while suppressing degradation interference.

\textbf{Results under Nonuniformity Conditions.} 
Table \ref{tab:nuc_results} reveals that while DMRN improves PSNR, its detection mAP drops to 35.24\%, indicating semantic feature loss during correction. Traditional methods (e.g., Liu, AHBC) also prove ineffective. In contrast, while UniCD shows strong performance (57.15\% mAP), DAISOD further elevates accuracy to 57.96\% with superior restoration fidelity (21.42 dB PSNR). This confirms that our strict physical constraints prevent the overfitting of high-frequency targets into the bias field, a common issue in prior works.

\begin{table}[t]
\centering
\caption{Quantitative comparison under Nonuniformity conditions. The best results are highlighted in bold.}
\label{tab:nuc_results}
\resizebox{\textwidth}{!}{
\begin{tabular}{lccccc}
\toprule
Strategy & Method & Pub. & PSNR $\uparrow$ & SSIM $\uparrow$ & mAP $\uparrow$ \\
\midrule
Direct & YOLO11n & - & 15.83 & 0.8033 & 38.64 \\
\midrule
\multirow{3}{*}{Separate} & Liu \cite{LIU2016235} + YOLO11n & IPT'16 & 18.37 & 0.8289 & 39.71 \\
 & DMRN \cite{chang2019infrared} + YOLO11n & GRSL'19 & 21.00 & 0.8456 & 35.24 \\
 & AHBC \cite{ahbc} + YOLO11n & TGRS'24 & 15.85 & 0.8067 & 38.57 \\
\midrule
Domain Adapt & SF-UT \cite{hao_simplifying_2024} & ECCV'24 & - & - & 46.92 \\
Union & UniCD \cite{fang2025detection} & CVPR'25 & 20.63 & 0.9085 & 57.15 \\
\midrule
\textbf{Ours} & \textbf{DAISOD} & - & \textbf{21.42} & \textbf{0.9132} & \textbf{57.96} \\
\bottomrule
\end{tabular}
}
\end{table}

\textbf{Qualitative Analysis.} 
As visualized in Fig. \ref{fig:detection_results}, we present visual comparisons with state-of-the-art methods under severe foggy and nonuniformity conditions. 
As can be seen, even in scenarios with heavy coupled degradations, our DAISOD achieves high-fidelity image restoration while accurately localizing small targets. 
This is because our physics-guided decoupling mechanism transforms the restoration into transparent parameter estimation, ensuring the effective removal of fog and bias without eroding weak target details. 
Meanwhile, the proposed contrastive domain alignment explicitly aligns the restored features with the clean domain, enhancing target saliency while suppressing background interference. 
In contrast, baseline methods like IA-YOLO \cite{liu2022image} and SF-UT \cite{hao_simplifying_2024} rely on opaque data-driven mappings or limited priors, which are prone to residual artifacts and missed detections in extreme environments.

\begin{figure}[!t]
  \centering
  \includegraphics[width=0.86\textwidth]{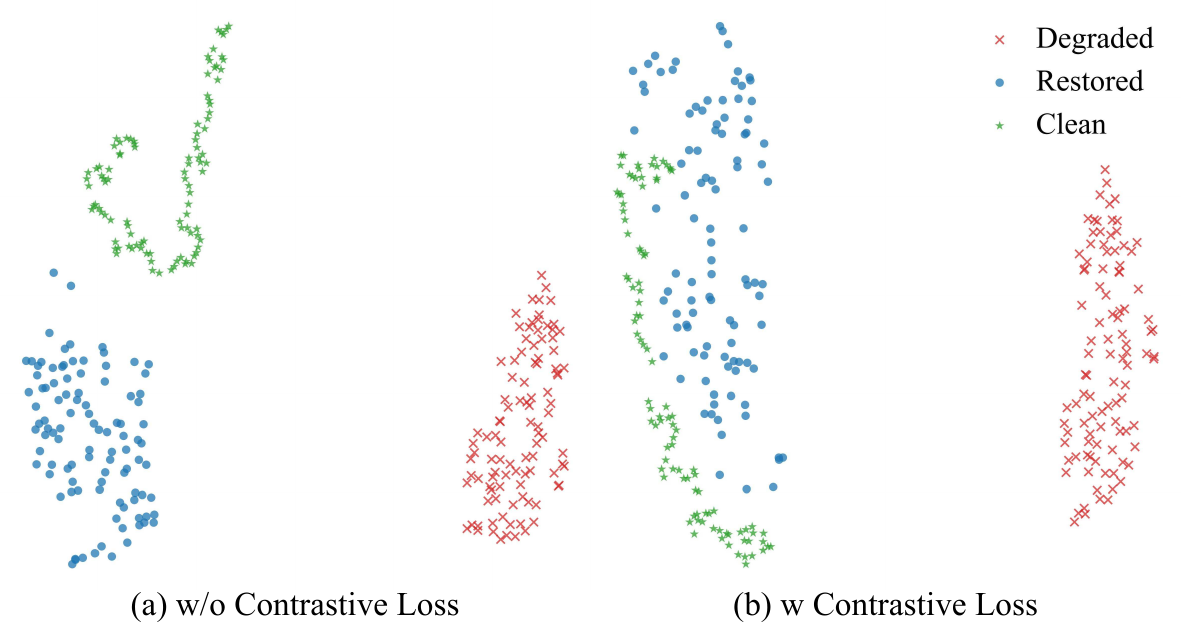} 

\caption{t-SNE visualization of feature distributions. 
(a) w/o Contrastive Loss: Restored features (Blue) are isolated from Clean features (Green), indicating a significant domain gap. 
(b) w/ Contrastive Loss: Restored features are pulled closer to the Clean cluster, effectively narrowing the domain gap. Meanwhile, their clear separation from Degraded features (Red) confirms degradation invariance.}
  \label{fig:tsne}
\end{figure}

\subsection{Ablation Study and Analysis}
To investigate the contribution of each component, we conduct a step-by-step ablation study on the IR-HDSOD dataset. The quantitative results are summarized in Table \ref{tab:ablation}.

\begin{table}[tbp]
\centering
\small
\setlength{\tabcolsep}{4pt}
\caption{Ablation study of the proposed DAISOD.}
\label{tab:ablation}
\resizebox{\textwidth}{!}{
\begin{tabular}{c|ccc|c}
\toprule
ID & \begin{tabular}[c]{@{}c@{}}Physics-Guided\\ Restoration \end{tabular} & \begin{tabular}[c]{@{}c@{}}Degradation-Aware\\ Fusion \end{tabular} & \begin{tabular}[c]{@{}c@{}}Contrastive Domain\\ Alignment \end{tabular} & mAP (\%) \\ \midrule
1  & -          & -          & -          & 34.54  \\
2  & \checkmark & -          & -          & 51.09  \\
3  & \checkmark & \checkmark & -          & 53.13  \\
4  & \checkmark & -          & \checkmark & 52.47  \\ 
5  & \checkmark & \checkmark & \checkmark & \textbf{53.97}  \\ \bottomrule
\end{tabular}
}
\end{table}

\textbf{Physics-Guided Restoration.} 
Compared with the baseline (Row 1), incorporating the physics-guided restoration module (Row 2) yields an absolute mAP gain of 16.55 percentage points. This substantial improvement demonstrates the importance of explicit physical modeling for separating coupled environmental and sensor degradations from vulnerable target signals. By transforming black-box restoration into transparent parameter estimation, this module removes fog and nonuniformity while preserving weak target details.

\textbf{Degradation-Aware Fusion.} 
Adding the degradation-aware fusion module to the physics-guided restoration model (Rows 2 and 3) improves mAP from 51.09\% to 53.13\%. Moreover, comparing Rows 4 and 5 shows that removing the fusion module while retaining physics-guided restoration and contrastive domain alignment reduces mAP from 53.97\% to 52.47\%. These comparisons demonstrate that adaptive feature selection reduces information redundancy and conflicting artifacts between parallel branches, thereby improving downstream representations.

\textbf{Contrastive Domain Alignment.} 
Finally, adding contrastive domain alignment to the model with physics-guided restoration and adaptive fusion (Rows 3 and 5) further improves mAP from 53.13\% to 53.97\%. By minimizing the feature-level contrastive loss, this mechanism successfully pulls the restored representations closer to the clean manifold while pushing them away from the degraded domain. This optimization effectively eliminates high-level degradation patterns in the latent space, ensuring the detector receives highly discriminative, degradation-invariant features.

\begin{figure}[!t]
	\centering
	\includegraphics[width=\textwidth]{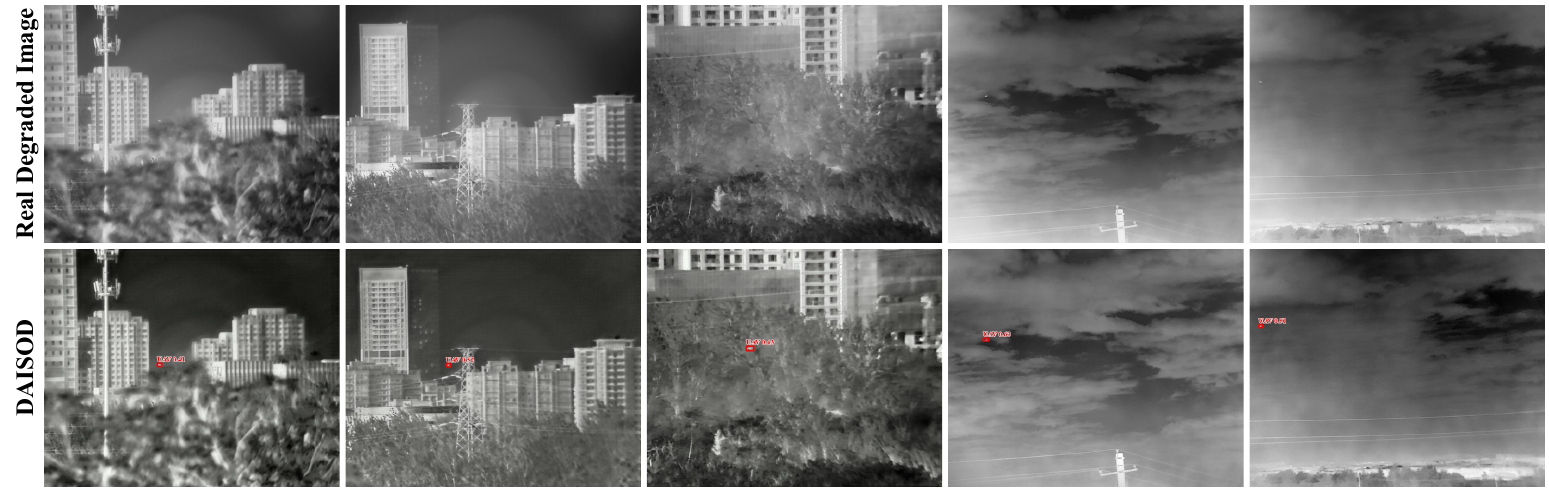}
	\caption{Restoration and detection results of DAISOD on real infrared images.}
	\label{fig:real_results}
\end{figure}

\textbf{Generalization to Real-World Infrared Images.}
To evaluate the sim-to-real generalization capability, we directly apply the model trained solely on the simulated IR-HDSOD training data to real-world infrared images, without any adaptation or fine-tuning. Since real degraded infrared images are scarce, and paired clean references and sufficiently comprehensive detection annotations are difficult to obtain, a statistically meaningful quantitative evaluation is currently infeasible. Therefore, we present qualitative restoration and detection results on a small collection of real infrared images gathered from publicly available online sources. As shown in Fig.~\ref{fig:real_results}, DAISOD effectively alleviates real-world degradations while preserving the overall scene structures and successfully localizing small targets across diverse urban and sky backgrounds. These qualitative results suggest that DAISOD generalizes favorably from simulated training data to real-world scenarios, providing qualitative evidence that the proposed synthetic dataset captures transferable degradation characteristics and that the proposed method remains effective under real-world conditions.

\textbf{Runtime Analysis.}
We further evaluate the computational efficiency of the complete DAISOD
pipeline on a separate deployment platform equipped with an Intel
Core i9-13980HX CPU and an NVIDIA RTX 4060 Laptop GPU with 8 GB memory
under Windows 11. The evaluation is conducted on 1,112 real infrared
images with an input resolution of $512\times512$. After GPU warmup, the
model-related computation, including physics-guided restoration,
adaptive fusion, YOLO11n detection, and non-maximum suppression, achieves
17.14 FPS. When image loading, preprocessing, bounding-box visualization,
and result saving are additionally included, the complete end-to-end
pipeline maintains 13.78 FPS. These results demonstrate that DAISOD
maintains practical computational efficiency despite introducing
multiple physics-guided restoration branches.

\section{Discussion and Future Work}
\label{sec:discussion}

The current study focuses on atmospheric fog and sensor nonuniformity, with the evaluation mainly covering aircraft and ship targets. Although the qualitative results on real infrared images indicate promising sim-to-real generalization, further evaluation on larger real-world datasets is still required. Moreover, the parallel restoration branches introduce additional computational overhead compared with a detector-only baseline.

Owing to its modular design, DAISOD can be extended to other degradation types without changing the overall framework. Specifically, for a new degradation, a corresponding restoration branch can be constructed according to its imaging characteristics. The degradation-aware estimator can then be extended to identify the new degradation, while the adaptive fusion module integrates the output of the added branch with the existing restoration results. Training samples covering the new degradation and different severity levels are also required to support this extension. For new target categories, the restoration and fusion modules can be retained because they mainly model degradation characteristics rather than target semantics. The framework can be adapted by adding annotated samples of the new targets and retraining or fine-tuning the detector. In future work, we will further extend DAISOD to support a broader range of degradation types and target categories.

\section{Conclusion}
\label{sec:conclusion}

To address the performance degradation and target erosion caused by adverse imaging conditions, this paper proposes DAISOD, a physics-guided framework for infrared small object detection. By explicitly estimating physical parameters and employing contrastive alignment, DAISOD achieves degradation-adapted restoration that effectively removes interference without damaging inherently weak target signals. Furthermore, we introduce the IR-HDSOD benchmark to facilitate comprehensive evaluation under various coupled degradation conditions. Extensive experimental results demonstrate that DAISOD consistently outperforms state-of-the-art methods, validating that the integration of explicit physical constraints is pivotal for balancing noise suppression and target preservation in complex infrared scenes.

\begin{credits}
	
	\subsubsection{\ackname}
	This work was supported in part by the National Key R\&D Program of China under Grant 2024YFB3909901; in part by the China Postdoctoral Science Foundation under Grant 2026M791648; in part by the National Natural Science Foundation of China under Grant U24B20139; and in part by the Hubei Provincial Natural Science Foundation of China under Grant 2026AFA040.
	
	\subsubsection{\discintname}
	The authors have no competing interests to declare that are relevant to the content of this article.
	
\end{credits}

\bibliographystyle{splncs04}
\bibliography{refs_daisod_complete}

@inproceedings{wang2023dehazing,
  title={Single image dehazing with deep-image-prior networks},
  author={H. Wang and X. Wang and Z. Su},
  booktitle={Proc. Int. Conf. Image Graph.},
  pages={78--90},
  year={2023}
}

@inproceedings{fu2025thermal,
  title={A lightweight and real-time asymmetric multi-output thermal radiation effects correction in infrared images},
  author={B. Fu and D. Xie and Y. Li and Y. Guo and X. Deng and Y. Zhang and Y. Shi},
  booktitle={Proc. Int. Conf. Image Graph.},
  pages={455--466},
  year={2025}
}

@inproceedings{zhang2025unsupervised,
  title={Unsupervised image restoration using domain discriminator with feature disentangle},
  author={C. Zhang and X. Ran and Z. Feng},
  booktitle={Proc. Int. Conf. Image Graph.},
  pages={3--15},
  year={2025}
}

@inproceedings{hu2025nighttime,
  title={Nighttime object detection with contextual auxiliary learning},
  author={X. Hu and X.-S. Zhang and Y.-J. Li and K.-F. Yang},
  booktitle={Proc. Int. Conf. Image Graph.},
  pages={231--242},
  year={2025}
}

@inproceedings{wang2025nonuniform,
  title={Non-uniform degradation aware and content complexity adaptive optimization for blind super-resolution},
  author={S. Wang and H. Shi and D. Xu},
  booktitle={Proc. Int. Conf. Image Graph.},
  pages={174--186},
  year={2025}
}

@inproceedings{tang2025ffta,
  title={{FFTA-Net}: A frequency-domain fusion and temporal alignment network for transmission line defect detection},
  author={H. Tang and J. Mao and Y. Wang and J. Yu and J. Yi and Z. He and Z. Tao and H. Zhang},
  booktitle={Proc. Int. Conf. Image Graph.},
  pages={116--129},
  year={2025}
}

@inproceedings{cheng2025hyperspectral,
  title={Hyperspectral image super-resolution via degradation-aware learning and frequency-domain feature enhancement},
  author={L. Cheng and J. Liu},
  booktitle={Proc. Int. Conf. Image Graph.},
  pages={218--230},
  year={2025}
}

@inproceedings{qin2020ffa,
  title={{FFA-Net}: Feature fusion attention network for single image dehazing},
  author={X. Qin and Z. Wang and Y. Bai and X. Xie and H. Jia},
  booktitle={Proc. AAAI Conf. Artif. Intell.},
  volume={34},
  pages={11908--11915},
  year={2020}
}

@inproceedings{zheng2023curricular,
  title={Curricular contrastive regularization for physics-aware single image dehazing},
  author={Y. Zheng and J. Zhan and S. He and J. Dong and Y. Du},
  booktitle={Proc. IEEE Conf. Comput. Vis. Pattern Recognit.},
  pages={5785--5794},
  year={2023}
}

@inproceedings{fang2025guided,
  title={Guided Real Image Dehazing Using {YCbCr} Color Space},
  author={W. Fang and J. Fan and Y. Zheng and J. Weng and Y. Tai and J. Li},
  booktitle={Proc. AAAI Conf. Artif. Intell.},
  volume={39},
  pages={2906--2914},
  year={2025}
}

@article{LIU2016235,
  title = {Correction of aeroheating-induced intensity nonuniformity in infrared images},
  journal = {Infrared Phys. Technol.},
  volume = {76},
  pages = {235--241},
  year = {2016},
  author = {L. Liu and L. Yan and H. Zhao and X. Dai and T. Zhang}
}

@article{chang2019infrared,
  title={Infrared aerothermal nonuniform correction via deep multiscale residual network},
  author={Y. Chang and L. Yan and L. Liu and H. Fang and S. Zhong},
  journal={IEEE Geosci. Remote Sens. Lett.},
  volume={16},
  number={7},
  pages={1120--1124},
  year={2019}
}

@article{ahbc,
  author={J. Xie and L. Song and H. Huang},
  journal={IEEE Trans. Geosci. Remote Sens.}, 
  title={Thermal Radiation Bias Correction for Infrared Images Using Huber Function-Based Loss}, 
  year={2024},
  volume={62},
  pages={1--15}
}

@inproceedings{liu2022image,
  title={Image-Adaptive {YOLO} for Object Detection in Adverse Weather Conditions},
  author={W. Liu and G. Ren and R. Yu and S. Guo and J. Zhu and L. Zhang},
  booktitle={Proc. AAAI Conf. Artif. Intell.},
  volume={36},
  pages={1792--1800},
  year={2022}
}

@inproceedings{hao_simplifying_2024,
  title = {Simplifying Source-Free Domain Adaptation for Object Detection: Effective Self-Training Strategies and Performance Insights},
  author = {Y. Hao and F. Forest and O. Fink},
  booktitle = {Proc. Eur. Conf. Comput. Vis.},
  year = {2024}
}

@inproceedings{fang2025detection,
  title={Detection-Friendly Nonuniformity Correction: A Union Framework for Infrared UAV Target Detection},
  author={H. Fang and X. Wang and Z. Li and L. Wang and Q. Li and Y. Chang and L. Yan},
  booktitle={Proc. IEEE Conf. Comput. Vis. Pattern Recognit.},
  pages={11898--11907},
  year={2025}
}

@article{dai2021attentional,
  title={Attentional local contrast networks for infrared small target detection},
  author={Y. Dai and Y. Wu and F. Zhou and K. Barnard},
  journal={IEEE Trans. Geosci. Remote Sens.},
  volume={59},
  number={11},
  pages={9813--9824},
  year={2021}
}

@article{narasimhan2002vision,
  title={Vision and the atmosphere},
  author={S. G. Narasimhan and S. K. Nayar},
  journal={Int. J. Comput. Vis.},
  volume={48},
  number={3},
  pages={233--254},
  year={2002}
}

@inproceedings{li2017aod,
  title={Aod-net: All-in-one dehazing network},
  author={B. Li and X. Peng and Z. Wang and J. Xu and D. Feng},
  booktitle={Proc. IEEE Int. Conf. Comput. Vis.},
  pages={4770--4778},
  year={2017}
}

@inproceedings{ying2023mapping,
  title={Mapping degeneration meets label evolution: Learning infrared small target detection with single point supervision},
  author={X. Ying and L. Liu and Y. Wang and R. Li and N. Chen and Z. Lin and W. Sheng and S. Zhou},
  booktitle={Proc. IEEE Conf. Comput. Vis. Pattern Recognit.},
  pages={15528--15538},
  year={2023}
}

@article{li2022dense,
  title={Dense nested attention network for infrared small target detection},
  author={B. Li and C. Xiao and L. Wang and Y. Wang and Z. Lin and M. Li and W. An and Y. Guo},
  journal={IEEE Trans. Image Process.},
  volume={32},
  pages={1745--1758},
  year={2022}
}

@inproceedings{li2022single,
  author={Z. Li and C. Zheng and H. Shu and S. Wu},
  booktitle={Proc. IEEE Int. Conf. Image Process.}, 
  title={Single Image Dehazing via Model-Based Deep-Learning}, 
  year={2022},
  pages={141--145}
}

@inproceedings{liu2023infrared,
  author={Z. Liu and J. He and Y. Zhang and T. Zhang and Z. Han and B. Liu},
  booktitle={Proc. IEEE Int. Conf. Image Process.}, 
  title={Infrared Small Target Detection Based on Saliency Guided Multi-Task Learning}, 
  year={2023},
  pages={3459--3463}
}

@inproceedings{joint_restoration,
  author={J. Ma and M. Lin and G. Zhou and Z. Jia},
  booktitle={Proc. IEEE Int. Conf. Image Process.}, 
  title={Joint Image Restoration For Domain Adaptive Object Detection In Foggy Weather Condition}, 
  year={2024},
  pages={542--548}
}

@inproceedings{domain-invariant,
  author={T. Kim and J. Na and J. Hwang and W. Hwang},
  booktitle={Proc. IEEE Int. Conf. Image Process.}, 
  title={Stay Focus on Object: Cross-Domain Detection Using Domain-Invariant Object Representation}, 
  year={2024},
  pages={2487--2493}
}

@inproceedings{itti2025bayesian,
  title={Bayesian Surprise for Small and Sub-Pixel Moving Target Detection},
  author={L. Itti and D. Liu and C. Teeter and S. Musuvathy},
  booktitle={Proc. IEEE Int. Conf. Image Process.},
  pages={815--820},
  year={2025}
}

@misc{yolo11_ultralytics,
  author = {G. Jocher and J. Qiu},
  title = {Ultralytics YOLO11},
  year = {2024},
  howpublished = {\url{https://github.com/ultralytics/ultralytics}}
}

@article{wu2022uiu,
  title={{UIU-Net}: {U-Net} in {U-Net} for infrared small object detection},
  author={X. Wu and D. Hong and J. Chanussot},
  journal={IEEE Trans. Image Process.},
  volume={32},
  pages={364--376},
  year={2022}
}

@inproceedings{wu2021contrastive,
	title={Contrastive learning for compact single image dehazing},
	author={H. Wu and Y. Qu and S. Lin and J. Zhou and R. Qiao and Z. Zhang and Y. Xie and L. Ma},
	booktitle={Proc. IEEE Conf. Comput. Vis. Pattern Recognit.},
	pages={10551--10560},
	year={2021}
}

@article{oord2018representation,
	title={Representation learning with contrastive predictive coding},
	author={A. Oord and Y. Li and O. Vinyals},
	journal={arXiv preprint arXiv:1807.03748},
	year={2018}
}

@inproceedings{schroff2015facenet,
	title={Facenet: A unified embedding for face recognition and clustering},
	author={F. Schroff and D. Kalenichenko and J. Philbin},
	booktitle={Proc. IEEE Conf. Comput. Vis. Pattern Recognit.},
	pages={815--823},
	year={2015}
}

\end{document}